\documentclass[runningheads]{llncs}
\usepackage{wrapfig}
\usepackage{graphicx}
\usepackage{cite}
\usepackage{amsmath,amssymb,amsfonts}
\usepackage{algorithmic}
\usepackage{textcomp}
\usepackage{xcolor}
\usepackage{tikz}
\usepackage{lipsum}
\usepackage{comment}
\usepackage{pgfplots}
\usepackage{subcaption}
\usepackage{siunitx}
\usepackage{url}
\usepackage{hyperref}
\usepackage{svg}
\usepackage{dblfloatfix}
\usepackage{soul}
\usepackage{booktabs}
\begin{document}
\title{
t-EVA: Time-Efficient t-SNE Video Annotation }
%
%\titlerunning{Abbreviated paper title}
% If the paper title is too long for the running head, you can set
% an abbreviated paper title here
%
\author{Soroosh Poorgholi \and
Osman Semih Kayhan \and
Jan C. van Gemert}
\authorrunning{S. Poorgholi et al.}
% First names are abbreviated in the running head.
% If there are more than two authors, 'et al.' is used.
%
\institute{Computer Vision Lab \\Delft University of Technology, Delft, The Netherlands \\
\email{s.poorgholi74@gmail.com}, \email{\{O.S.Kayhan, J.C.vanGemert\}@tudelft.nl}}
% }
% \\
% \url{http://www.springer.com/gp/computer-science/lncs} \and
% ABC Institute, Rupert-Karls-University Heidelberg, Heidelberg, Germany\\
% \email{\{abc,lncs\}@uni-heidelberg.de}}
%
\maketitle              % typeset the header of the contribution
\begin{abstract}
Video understanding has received more attention in the past few years due to the availability of several large-scale video datasets. 
However, annotating large-scale video datasets are cost-intensive. In this work, we propose a time-efficient video annotation method using spatio-temporal feature similarity and t-SNE dimensionality reduction to speed up the annotation process massively.
Placing the same actions from different videos near each other in the two-dimensional space based on feature similarity helps the annotator to group-label video clips.
We evaluate our method on two subsets of the ActivityNet (v1.3) and a subset of the Sports-1M dataset. We show that t-EVA\footnote{\url{https://github.com/spoorgholi74/t-EVA}} can outperform other video annotation tools while maintaining test accuracy on video classification. 
%The code will be publicly available in the t-EVA GitHub repository \footnote{\url{https://github.com/spoorgholi74/t-EVA}}.

\end{abstract}
\section{\textbf{Introduction}}

The availability of large-scale video datasets \cite{activitynet-refrence,kinetics-DBLP:journals/corr/KayCSZHVVGBNSZ17,sport1mCVPR14} has made video understanding in various tasks such as action recognition \cite{t-c3d-Liu2018TC3DTC,C3D-DBLP:journals/corr/TranBFTP14,TSN-DBLP:journals/corr/WangXW0LTG16}, object tracking \cite{tracking1-zhang2020simple,tracking2-DBLP:journals/corr/abs-1812-05050,tracking3-DBLP:journals/corr/abs-1808-06048} an attractive topic of research.
Various supervised methods \cite{C3D-DBLP:journals/corr/TranBFTP14,T3D-DBLP:journals/corr/abs-1711-08200,TSN-DBLP:journals/corr/WangXW0LTG16} have improved video classification and temporal localization accuracy on large-scale video datasets such as ActivityNet (v1.3) \cite{activitynet-refrence}; however, labeling videos on such a large-scale dataset, requires a great deal of human effort. 
Therefore, other methods aim to train the networks for tasks such as video action recognition in a semi-supervised \cite{semisupervised1-DBLP:journals/corr/abs-1801-07230,semisupervised2-DBLP:journals/corr/abs-1801-07827} manner without having the full labels.
To decrease the dependency on the quality and amount of annotated data, \cite{weakly1-DBLP:journals/corr/abs-1901-09244,weakly2-DBLP:journals/corr/abs-1905-00561} investigate pre-training features with internet videos with noisy labels in a weakly supervised manner. 
% Self-supervised methods have also been proposed \cite{self-supervised1-DBLP:journals/corr/FernandoBGG16}, where the frame order in unlabeled videos can help the learning process.
However, these methods do not achieve higher accuracy on video classification tasks than supervised models on large-scale video datasets such as Kinetics \cite{kinetics-DBLP:journals/corr/KayCSZHVVGBNSZ17}. Instead of using such techniques, we focus on reducing the annotation effort for adding more training data.

\begin{comment}
%%%%%%%%%%%%%%%%%%%%%%%%%%%%%%%%%%
\begin{figure}
\centering
\includegraphics[width=\columnwidth]{Figures/tsne_10class_400vid.png}
\caption{t-SNE projection of extracted features from 407 ActivityNet (v1.3) videos sampled at 128 frames per clip \textcolor{red}{Fig. 1 is the key image of the paper. What do you want to sell here? How can you sell it the paper by it (how can you improve?}}
\label{fig:tsne400}
\end{figure}
%%%%%%%%%%%%%%%%%%%%%%%%%%%%%%%%%
\end{comment}

Fully-supervised models require much annotated data that is unavailable as videos are unlabeled by nature, and annotating them is labor-intensive. 
Large scale datasets \cite{kinetics-DBLP:journals/corr/KayCSZHVVGBNSZ17, activitynet-refrence,epic-kitchen-DBLP:journals/corr/abs-1804-02748} use strategies like \textit{Amazon Mechanical Turk} (AMT) to annotate the videos. \cite{kinetics-DBLP:journals/corr/KayCSZHVVGBNSZ17} uses majority voting between multiple AMT workers to accept annotation of a single video. Using such methods is not efficient for video annotation on a large scale as it costs a lot in terms of time and money. 
MuViLab~\cite{muvilab-github}, an open-source software, enables the oracle to annotate multiple parts of a video simultaneously.
%Other tools have been proposed \cite{muvilab-github} to make the annotation process easier by providing open-source software such as \textit{MuViLab} that enables the oracle to annotate multiple parts of a video at the same time. 
However, these methods do not exploit the structure of the video data.

%%%%%%%%%%%%%%%%%%%%%%%%%%%%%%%%%%

\begin{wrapfigure}{R}{0.6\textwidth}
  \begin{center}
    \includegraphics[width=0.6\textwidth]{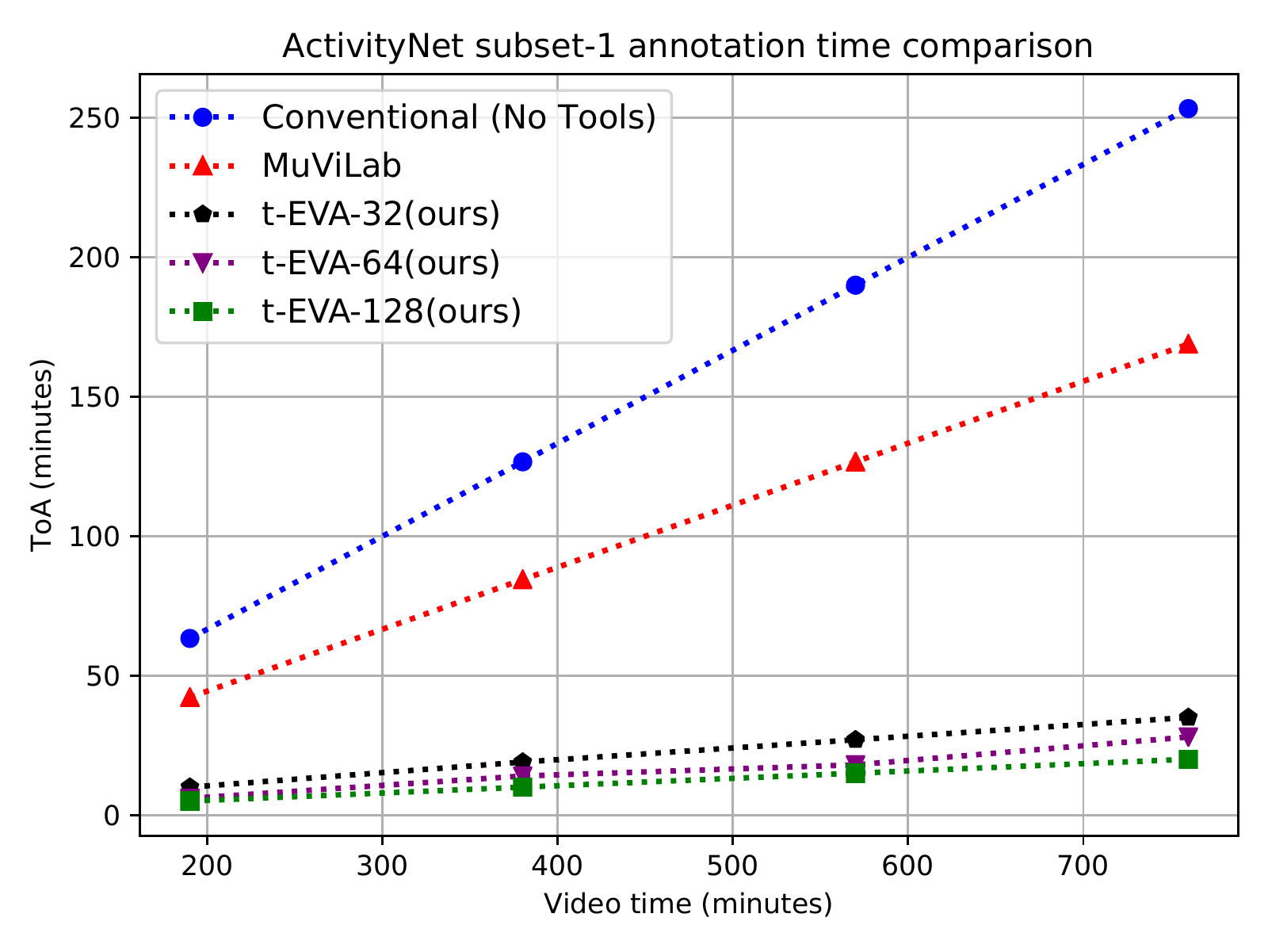}
  \end{center}
  \caption{Comparison of annotation time using different tools versus video time for the ActivityNet \cite{activitynet-refrence} subset-1. Our annotation method (t-EVA) outperforms the conventional (no specific tools) annotation and MuViLab \cite{muvilab-github} in annotation time. With a window size of 128 time-steps (128-TS), our method can annotate 769 minutes of video in 21 minutes. The MuViLab and conventional annotation numbers are extrapolated.}
\label{fig:subset1-toa}
\end{wrapfigure}

%%%%%%%%%%%%%%%%%%%%%%%%%%%%%%%%%

%In this work, 
We introduce an annotation tool that helps the annotator group-label videos based on their latent space feature similarity in a 2-dimensional space.
Transferring the high-dimensional features obtained from 3D ConvNet to two dimensions using t-SNE gives the annotator an easy view to group label the videos both, temporal labels and classification labels. The annotation speed depends on the quality of the extracted features and how well they are placed together in the t-SNE plot. 
If the classes are well-separated in the t-SNE plot, group labeling becomes faster for the oracle.

We evaluate our method on two subsets of ActivityNet (v1.3 datasets)\cite{activitynet-refrence} and a subset of Sports-1M dataset \cite{sport1mCVPR14} with 15 random classes.
\textit{Conventional annotation} refers to humans watching the videos and annotating the temporal boundaries of the human actions in videos without any specific tool. \textit{MuViLab} is a more advanced open-source tool that extracts short clips from each video and plays them simultaneously in a grid-like figure beside each other. Oracle can annotate the video by selecting multiple short clips at the same time and assigning the specific class.
We show that t-EVA outperforms conventional annotation techniques (with no specific tools) and 
%a open-source software called 
MuViLab \cite{muvilab-github} in time of annotation (ToA) by a large margin on the ActivityNet dataset while still being able to keep the test accuracy on video classification task within a close range of using the original ground truth annotations (Figure~\ref{fig:subset1-toa}). 

% \begin{comment}
% Our contributions in this paper are the following: (i) A smart annotation tool to make video annotation more time-efficient \textcolor{red}{what can be more contribution?}

% showing how incremental annotation and training can help annotation by separating the features better (make it more precise at the end)
% Ablation study on the choice of parameters and methods.
% We also discuss the limitation of our method in labeling complex actions of the Breakfast dataset.
% \hl{As another contribution, we adapt the method used in XXX and YYY (MLP) to exploit the temporal information in the sequential actions such as the Breakfast dataset and show the result in section XXX.}
% \end{comment}

\section{\textbf{Related Work}}
%\subsection{Video understanding}
\begin{comment}
Goal: different supervised learning methods has been applied to video understanding and action recognition in video (single stream, two stream). They achieved the accuracy of xxx but unsupervised (ref) and semi-supervised methods was not able to reach the same accuracy.

Keywords: hand crafted features, Supervised, C3D, t3d, TSN, single stream, two stream, optical flow, Unsupervised methods, semi supervised methods, classification accuracy, action recognition.

conclusion: Supervised methods works the best and need labels
\end{comment}
\textbf{Video Understanding.} In the past, the focus was on the use of specific hand-designed features such as HOG3D \cite{hog3d-Klser2008ASD} SIFT-3D \cite{sift3d-10.1145/1291233.1291311}, optical flow \cite{optical-flow-DBLP:journals/corr/abs-1712-08416} and iDT \cite{iDT-6751553}.  Among these methods, iDT and Optical flow is being used in combination with CNNs in different architectures such as two-stream networks \cite{two-stream-DBLP:journals/corr/SimonyanZ14}. Afterwards, some methods use 2D CNNs and extract features from video frames and combine them with different temporal integration functions \cite{disc-cnn-DBLP:journals/corr/XuYH14, action-vlad-Girdhar-2017-113316}.
% \cite{fusion-DBLP:journals/corr/abs-1806-03863} gives an analysis of how to integrate temporal information of video frames with different fusion methods ( "\textit{late fusion}", "\textit{slow fusion}", etc.) by using additional temporal convolutions on top of the spatial convolutions. 
The introduction of 3D convolutional \cite{multi-stage-DBLP:journals/corr/ShouWC16, C3D-DBLP:journals/corr/TranBFTP14} in CNNs which extend the 2D CNNs in temporal dimension show promising results in the task of action recognition in large-scale video datasets. 
3D CNNs in different variations such as single stream and multiple-stream are among state of the art in the task of video understanding \cite{quo-vado-DBLP:journals/corr/CarreiraZ17,slow-fast-Feichtenhofer_2019_ICCV,vid-transformer-DBLP:journals/corr/abs-1812-02707,timeception-DBLP:journals/corr/abs-1812-01289,disc-filter-martinez2019action, spatio-temoral-DBLP:journals/corr/abs-1711-10305, channel-seperated-DBLP:journals/corr/abs-1904-02811}. In this paper, we utilize single a stream 3D CNN architecture to obtain video features.

\textbf{Dimensionality Reduction.}
Dimensionality reduction (DR) is an essential tool for high-dimensional data analysis. In linear DR methods such as PCA, the lower-dimension representation is a linear combination of the high-dimensional axes. Non-linear methods, on the other hand, are more useful to capture a more complex high-dimensional pattern \cite{non-linear-book}. In general, non-linear DR tries to maintain the local structure of the data in the transition from high-dimension to low-dimension and tends to ignore larger distances between the features \cite{tivi-sne-Chatzimparmpas_2020}. 
t-Distributed Stochastic Neighbor Embedding (t-SNE) introduced by \cite{tsne-vandermaten-7b54165e73a3424b8820136bcf61ca89} is a non-linear DR technique which is used more for visualization. 
% However, t-SNE has raised some concerns regarding the reliability and interpret-ability of the results. \cite{tsne-pitfall-Wattenberg2016HowTU} mentions some of the disadvantages of using t-SNE such as 1) reliability of the results on the hyper-parameter choice
% 2) The fact that the cluster size might not mean anything 3) Distance between clusters might be deceiving. However, the data used in \cite{tsne-pitfall-Wattenberg2016HowTU} is artificial data in which the distribution is known before applying t-SNE.
\cite{tsne-prov-DBLP:journals/corr/LindermanS17} shows that t-SNE is able to distinct well-separable clusters in low-dimensional space. Moreover, some works \cite{tivi-sne-Chatzimparmpas_2020, UMAP-lel2018umap, gpgpu-PTMHLLEV20} propose for more effective use of t-SNE . \cite{tivi-sne-Chatzimparmpas_2020} proposes a tool to support interactive exploration and visualization of high-dimensional data. An alternative to t-SNE is UMAP  \cite{UMAP-lel2018umap}; however, t-SNE is well studied, shows good results, and has the benefit of high-speed optimization \cite{gpgpu-PTMHLLEV20}. t-EVA uses t-SNE to reduce the dimensionality of the feature representations.

\textbf{Data Annotation} is essential for supervised models. Different tools are proposed for making an easy annotation tool for videos and images. However, they usually do not exploit the structure of the data, which is especially useful in videos \cite{muvilab-github, labelImg-github, imglab-github}.
Some methods \cite{internet-annot-4420087, multi-instance-5459194, DML-Wu_2015_CVPR, AL-tensorboardDBLP:journals/corr/abs-1901-00675} are designed to make the process of image annotation easier. \cite{internet-annot-4420087} offers a real-time framework for annotating internet images, and \cite{multi-instance-5459194} uses multi-instances learning to learn the classes and image attributes together; however, none of these methods use a deep representation of data. In more recent works, \cite{DML-Wu_2015_CVPR} utilizes \textit{Deep Multiple Instance Learning} to automatically annotate images and \cite{AL-tensorboardDBLP:journals/corr/abs-1901-00675} uses semi-supervised t-SNE and feature space visualization in lower dimension to provide an interactive annotation environment for images. \cite{VIA-10.1145/3343031.3350535} proposes a general framework for annotating images and videos. However, to the best of our knowledge, our method is the first video annotation platform that can \textit{exploit the structure} of video using latent space feature similarity to increase the annotation speed.
%%%%%%%%%%%%%%%%%%%%%%%%%%%%%%%%%%
\begin{figure*}[ht]
\centering
\includegraphics[width=\textwidth]{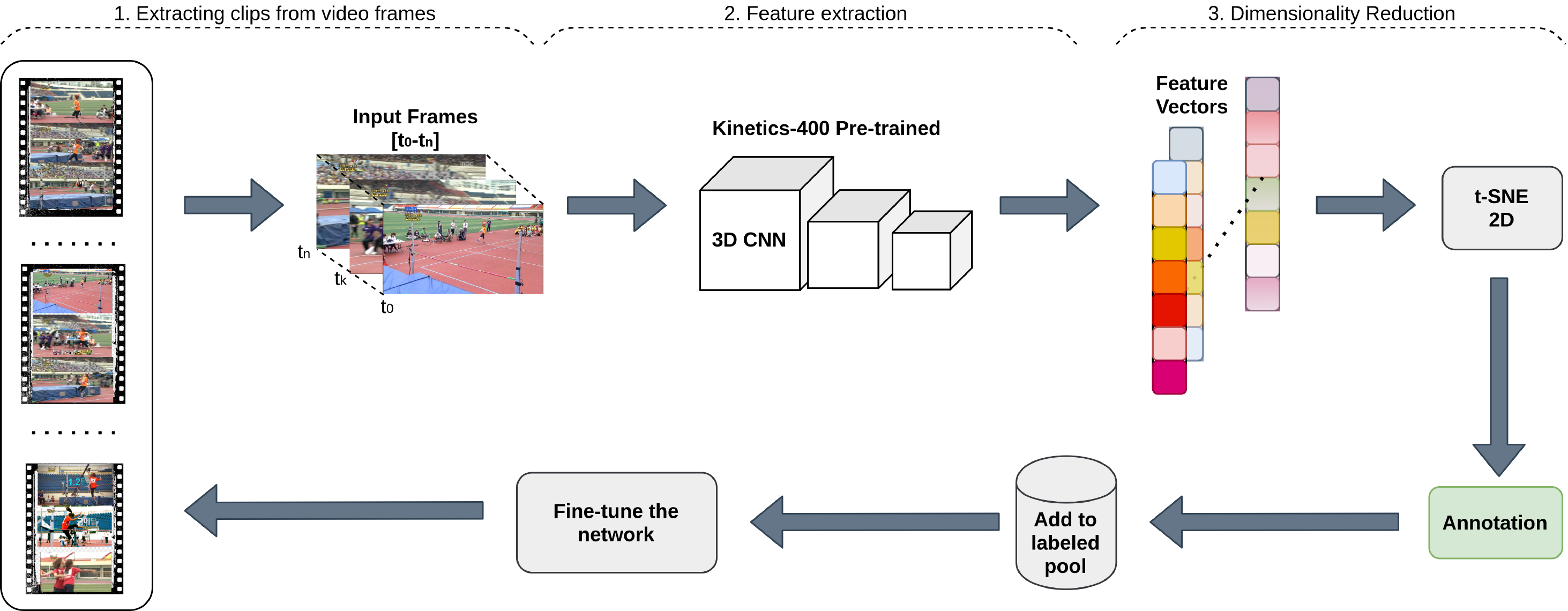}
\caption{t-EVA pipeline: 1) Video clips are extracted from \textit{n} consecutive frames [$t_0$-$t_n$] (time-steps). 2) Spatio-temporal features are extracted from the last layer of a 3D ConvNet before the classifier layer. 3) High dimensional features are projected to two dimensions using t-SNE and are plotted on a scatter plot. 4) Oracle annotates the clips represented in the scatter plot using a lasso tool. 5) The newly annotated data is added to the labeled pool. 6) The network is fine-tuned for a certain number of epochs. This cycle is repeated until all the videos are labeled, or the annotation budget runs out.}
\label{fig:pipeline}
\end{figure*}
%%%%%%%%%%%%%%%%%%%%%%%%%%%%%%%%%

\section{\textbf{t-EVA for efficient video annotation}}
%\subsection{\textbf{Pipeline}}
%%%%%%%%%%%%%%%%%%%%%%%%
\begin{comment}
Goal: explain the general pipeline. Introduce that the different parts will be described in more details in the following sections.In summary the pipeline include the following steps 1. 2. 3. \\

keywords: Tsne, C3d, features, temporal enhancement, dimensionality reduction, interactive labeling matplotlib.
\end{comment}
%%%%%%%%%%%%%%%%%%%%%%%%

\begin{figure*}[ht]
\centering
\includegraphics[width=\textwidth]{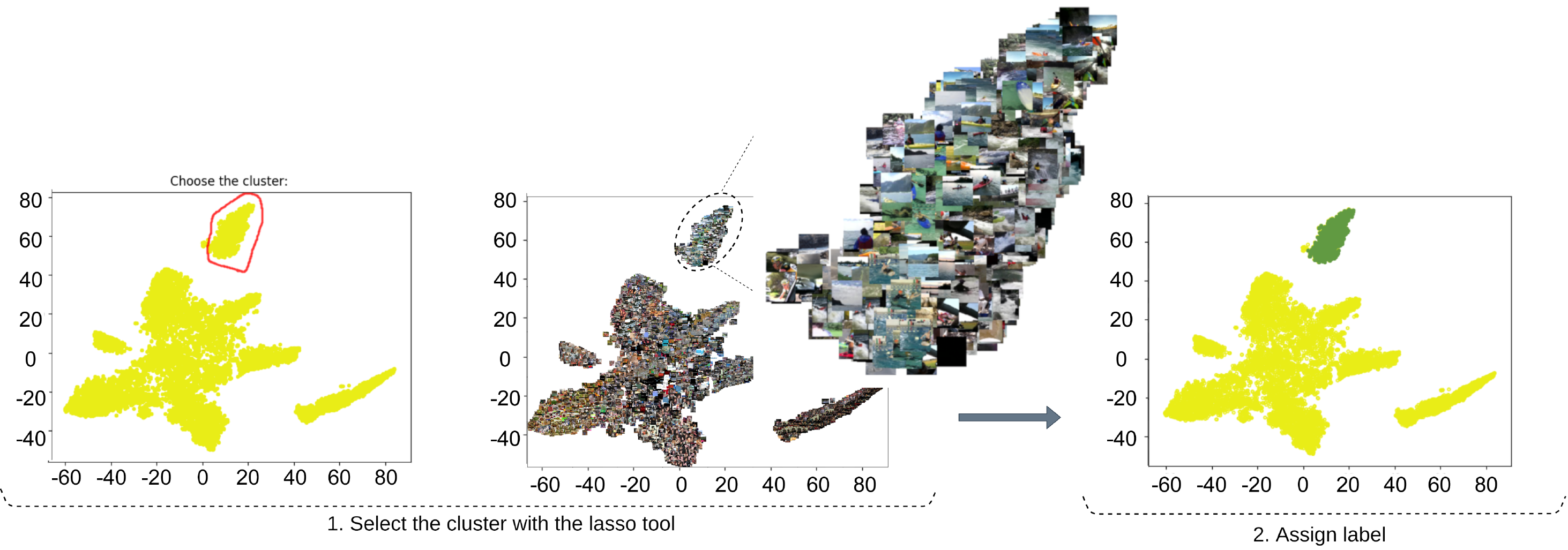}
\caption{A minimal representation of the annotation tool. 1) The oracle can see the scatter plot (left) and the corresponding frames from the videos (middle) in separate figures. 2) Based on the figures' inspection, the oracle can detect different clusters of an action class (kayaking) and use the lasso tool to select the cluster. 3) In the end, the oracle assigns a  label and based on the assigned class name, the selected points in the scatter plot change color.}
\label{fig:annotation}
\end{figure*}

We propose incremental labeling with t-SNE based on feature similarity (Figure~\ref{fig:pipeline}). First, several videos are randomly selected from the unlabeled pool, and 3D ConvNet features are extracted. The feature embeddings are transferred to a two-dimensional space using t-SNE.
As it can be seen in Figure~\ref{fig:annotation}, the oracle has two subplots for annotation: (i) A plot in which the oracle can use a lasso tool to group label videos and (ii) Other plot with the middle frame of each clip in which the oracle can move and zoom with the cursor on the plot and observe where to annotate.
After annotating the first set of videos, the video clips are moved to the labeled pool, and the 3D network is fine-tuned for a certain number of epochs with the newly labeled videos. We continue this process until all the videos are labeled, or the annotation budget finishes. 

%%%%%%%%%%%%%%%%%%%%%%%%%%%%%%%%%

%\subsection{\textbf{Dimensionality Reduction}}
\begin{comment}
Goal: Why TSNE dimensionality reduction has been used? Why Barnes-Hut version and what is the parameters? how this help the labeling for the oracle.\\

Keywords: Tsne, PCA, Linear and non-linear, 512 to 2 dimension, will cause loss of information, how can this be compensate? 
\end{comment}

We use 3D ConvNets to extract features from the videos and split each video \textit{v} into \textit{k} shorter clips $v_i=[clip_1, ..., clip_k]$ by sampling every \textit{n} non-overlapping frames $clip_i=[frame_1, ..., frame_n]$. Sampling in multiple time-steps enables us to capture different lengths of actions in the dataset. Afterwards, each clip \textit{$c_i$} is fed into the 3D ConvNet, for feature extraction. The features are extracted from the last convolution layer after applying global average pooling. In t-SNE, the pair-wise distances between feature vectors are used to map features to 2D.
In this paper, we use the Barnes-Hut optimized t-SNE version~\cite{barnes-haut-maaten2013barneshutsne}, which reduces the complexity of $O(N log N)$ where \textit{N} is the number of data-points.

\subsection{\textbf{How to Annotate?}}
\begin{comment}
Goal: explain how the annotation tool works
keywords: matplotlib, lasso tool, spatio-temporal features, feature similarity, incremental annotation
\end{comment}
%%
An overview of the annotation procedure can be seen in Figure~\ref{fig:annotation}.
%shows an overview of the annotation tool, which works in the following way. 
First, the oracle sees the scatter plot with all points with the same color representing the unlabeled pool (Figure~\ref{fig:annotation} left) and the corresponding middle frame of each clip in the video (Figure~\ref{fig:annotation} middle).
The oracle can move the cursor and zoom in the plot to inspect the frames with more details.
Second, using the lasso tool, the oracle can draw a lasso around the scatter plots based on the visual similarity and inspection of the video frames. 
Third, oracle assigns the labels, and the network is fine-tuned for a certain number of epochs. 
The same process repeats until all the videos are annotated, or the annotation budget ends. 
%%%%%%%%%%%%%%%%%%%%%%%%%%%%%%%

\begin{comment}
Goal: explain how the training is done. What are the training parameters? how incremental training and labeling can help. (maybe mention in the experiment)
\end{comment}

\section{\textbf{Experiments}}
In this section, we first explain the benchmark dataset and evaluation metrics. In addition, we empirically show how our t-EVA method can speed up annotation for the ActivityNet dataset while keeping the video classification accuracy in a close range to the usage of the ground truth labels. We also compare our results with MuViLab \cite{muvilab-github} annotation tool. 
Furthermore, we qualitatively show how t-EVA can help to annotate the Sports1-M \cite{sport1mCVPR14}.

\subsection{\textbf{Datasets}}
\textbf{ActivityNet (v1.3)} is an untrimmed video dataset with a wide range of human activities \cite{activitynet-refrence}. It comprises of 203 classes with an average of 137 untrimmed videos per class in about 849 hours of video. We use two subsets of the ActivityNet dataset.
The first subset comprises 10 random classes, namely \textit{preparing salad, kayaking, fixing bicycle, mixing drinks, bathing dog, getting a haircut, snatch, installing carpet, hopscotch, zumba} consisting of 607 videos with 407 training videos and 200 testing videos. The second subset adds another 5 handpicked classes, which are \textit{playing water polo, high jump, discus throw, rock climbing, using parallel bars}, and they are visually close to some of the 10 random classes to make the classification task harder. The second subset comprises 950 videos with 639 videos in training and 311 videos in the test set.

\textbf{Sports-1M} is a large-scale public video dataset with 1.1 million YouTube videos of 487 fine-grained sports classes \cite{sport1mCVPR14}. We choose a subset of 15 random classes of the Sports-1M dataset, namely \textit{boxing, kyūdō, rings (gymnastics), yoga, judo, skiing, dachshund racing, snooker, drag racing, olympic weightlifting, motocross, team handball, hockey, paintball, beach soccer} with 702 videos in total. The dataset provides video level annotation for the entire untrimmed video; however, the temporal boundaries of the actions in the video are not identified. Approximately 5\% of the videos contain more than one action label.

% \textbf{Breakfast} is a dataset for human cooking activities from multiple cameras with multiple view-points \cite{breakfast-refrence}. It includes 1712 videos with an average length of 2.3 minutes per video. It contains 10 breakfast preparation classes from 18 different kitchens. Each video represents only one activity class, but it contains sub-actions such as \textit{"put oil in pan", "cracking egg", "frying egg", "putting egg in dish"} under "\textit{fried egg}". The creators of the dataset have provided temporal boundary annotation for each sub-action. 

\subsection{\textbf{Evaluation Metrics}}
\begin{comment}
goal: explain how do we measure our superiority in different settings:
- explain what have been used for each dataset
- dont make it super general

\end{comment}

To evaluate our method on ActivityNet subsets, we report the \textit{time of annotation} (ToA) as a metric to measure how fast the oracle can annotate a certain number of videos. The ToA score is an average of three times repeating each experiment by the oracle. ToA for conventional annotation and MuViLab on ActivityNet subset-1 is extrapolated since annotating 13 hours of video using these methods is not feasible. We also report video classification accuracy in the form of mean average precision (mAP) for the ActivityNet subsets to measure the quality of annotation when the network is fine-tuned with our annotations versus with the ground truth annotations. mAP is used instead of a confusion matrix since some videos of ActivityNet contain more than one action \cite{activitynet-refrence}.

For the Sports-1M \cite{sport1mCVPR14} dataset, we perform a qualitative analysis of the t-SNE projections. To motivate our design choices beyond qualitative results, we introduce a realistic annotation emulation metric to estimate the quality of t-SNE projections on a global and local level. 
To report how well the t-SNE projection can separate the classes at a global level, we use a measure of cluster homogeneity, and completeness. Homogeneity measures if the points in a cluster only belong to one class and completeness measures if all points from one class are grouped in the same cluster. 
In an ideal t-SNE projection, all the points in each cluster belong to one class (homogeneity=1.0), and all the points from a class are in the same cluster (completeness=1.0), which makes the annotation process much faster. For clustering, K-Means clustering with K being the number of classes is used. We use the K-Means clustering algorithm because it is fast and has less hyperparameters to choose.

Since ToA can be a subjective metric, to evaluate the generalization of t-EVA and to emulate the oracle's annotation speed better, we also use a measure of local homogeneity using K-nearest neighbors (KNN) with \textit{K=4} as in \cite{AL-tensorboardDBLP:journals/corr/abs-1901-00675}.

KNN can be used to estimate the local homogeneity between the features in lower dimensions. Higher  KNN accuracy results in higher local homogeneity and better grouping; namely, the oracle can annotate the videos faster.

\subsection{\textbf{Implementation Details}}
\begin{comment}
Goal: explain what architectures has been use
- how the network has been trained 
- where the features was extracted 
- the mlp and training
\end{comment}
\textbf{Feature Extraction.} We use the 3D ResNet-34 architecture~\cite{resnet-3d-DBLP:journals/corr/abs-1708-07632}, pre-trained on Kinetics-400, as a feature extractor for all the experiments owing to their good performance and usage of RGB frames only.
As in \cite{resnet-3d-DBLP:journals/corr/abs-1708-07632}, each frame is resized spatially to 112$\times$112 pixels from the original resolution. 
Each video is transferred to clips by sampling every 32 consecutive frames. 
The feature extractor in every forward pass takes a clip in the form of a 5D tensor as an input. Each dimension of the input tensor represents the batch size, input color channels, number of frames, spatial height, and width, respectively. Namely, an input tensor for a clip sampled at 32 frames can be shown as (1, 3, 32, 112, 112). 
The features are extracted after the final 3D average pooling with an 8x4x4 kernel before the classifier layer.
%(fully connected layers and softmax). 
The dimensions of the feature vectors are \textit{k}$\times$512 with \textit{k} being the total number of clips and later reduced to \textit{k}$\times$2 using t-SNE.

\textbf{t-SNE.} For dimensionality reduction, a Barnes-Hut implementation of t-SNE with two components are used from the scikit-learn library \cite{scikit-learn}. The perplexity is set to 30, and the early exaggeration parameter is 12, with a learning rate of 200. The cost function is optimized for 2500 iterations.

\textbf{Training.} After annotating each set of videos, the network is fine-tuned for a certain number of epochs. For training, the same 3D ResNet-34~\cite{resnet-3d-DBLP:journals/corr/abs-1708-07632} architecture is used.
The sample duration is chosen as 32 frames for each clip, and the input batch size is 32. 
%The 3D ConvNet takes a 5D tensor as an input. The parameters are the same as in the feature extractor.
Stochastic gradient descends (SGD) is used as the optimizer with a learning rate of 0.1, weight decay of 1\textit{e}-3, and momentum of 0.9.

% \textbf{Temporal Enhancement.} For temporal enhancement of the features, we use the ResNet-34 features of a single complex action to train a two-layer MLP to learn the sub-action coherency. For optimization, Adam optimizer, with a learning rate of 1\textit{e}-3, is used. After training the network for 100 epochs on features belonging to one complex action, we extract the features from the second layer of the MLP.

%%
\subsection{\textbf{Results on ActivityNet}}
\begin{comment}
goal: - explain what are we gonna do here.
- say what are the test settings 
- show the results
- draw the conclusion that our method can annotate activity-net subset 1 in faster time without loosing that much mAP
\end{comment}

\textbf{ActivityNet Subset-1.} First, we put all the 407 videos in the unlabeled pool. Then, we divide the videos randomly into four different sets of unlabeled videos. The clips are generated with 32 consecutive frames, and the features are extracted using the 3D Resnet-34. After annotating each set of unlabeled videos, the network is fine-tuned for 20 epochs with the labeled videos. To note that, previously labeled videos are also used in the later epochs. The process continues until the network reaches 100 epochs. Between epoch 60 and 100, the network is fine-tuned using all 407 videos. Meanwhile, we refine the labels of the videos. 

The videos are annotated incrementally, each time one set is labeled. Table \ref{tbl:incremental-annotation} shows that the annotation time drops after every iteration of annotation and fine-tuning.
Before fine-tuning the network, the labeling of the first set takes 600 seconds. ToA reduces 150 seconds at epoch 60 when the network is fine-tuned with previously labeled videos.
Because of the incremental labeling and fine-tuning, the network learns to extract better features from the videos, which can be better grouped in the t-SNE plot.
It is also expected that the oracle spends more time annotating the first few unlabeled set as the network is not yet fine-tuned. The quality of annotation at the early stage significantly impacts the next iterations of extracted features.

% Please add the following required packages to your document preamble:
% \usepackage{booktabs}
\begin{table}[h!]
\centering
\caption{Oracle's time of annotation (ToA) is shown on subset 1 of the ActivityNet (v1.3) dataset with 10 classes containing 407 videos ($\sim$13 hours). At every 20 iterations from 0 to 60, 102 new videos are annotated, and the network is fine-tuned for 20 epochs. From epoch 60 to 100, no new video is added. The previous video labels are refined by the oracle as the network can extract better features. The network is fine-tuned on the existing labeled videos until epoch 100. It can be seen with incremental annotation and fine-tuning the annotation time in the later epochs drops.}
\begin{tabular*}{\columnwidth}{@{\extracolsep{\fill}}ccccccc@{}}
\toprule
\textbf{Epoch}      & 0   & 20  & 40  & 60  & 80  & 100 \\ \midrule
\textbf{ToA (seconds)} & 600 & 552 & 516 & 450 & 240 & 180 \\ \bottomrule
\end{tabular*}
\label{tbl:incremental-annotation}
\end{table}

\begin{comment}
Goal: show how much our method increase time efficiency while being able to keep the same test accuracy
explain: 
\begin{itemize}
    \item how much gain in time you gain? 
    \item what are the conventional methods and how do they work? 
    \item how does incremental annotation helps the quality of annotation?
    \item How the clusters start separating and it get easier to label and the homogeneity increases.
\end{itemize}
\end{comment}

\textbf{Annotation Speed.} 
To evaluate the annotation speed, we choose three methods: conventional, MuViLab~\cite{muvilab-github}, and t-EVA. 

One way to increase the annotation speed of t-EVA is by putting more videos on the screen for the oracle to annotate. However, it does not make the labelling process easier. Since ActivityNet videos on average have 30 frames per second (FPS), every 32 time-steps that we sample represent almost 1 second $(\sim \frac{32}{30})$ of video. Putting all of the 407 videos (13 hours) overflows the screen with the frames and makes the annotation harder for the oracle. One way to prevent overflowing the figures with thousands of frames is to increase the time-steps for sampling frames from each clip to the point that the network can still preserve the clips' temporal coherency. This way, we can show all of the videos on the 2D plot with fewer points. Consequently, we design three different t-EVA in terms of the number of time steps as t-EVA-32, 64, and 128.

% We manually labeled those videos with MuViLab, conventional method and three t-EVA versions.
%\textcolor{red}{Since annotating the entire subset were not feasible for us, we estimated the time of annotation using MuViLab and Conventional annotation. We annotated 30 minutes of videos using these methods and extrapolated the result to match the total duration of first subset of ActivityNet. Later the entire subset-1 is annotated using different variations of t-EVA and we compare the annotation speed of all these methods (Table \ref{tbl:time-comp}).}
\begin{table}[h!]
\centering
\caption{Comparison of time gain when annotating with different methods on a subset-1 of ActivityNet containing 769 minutes of video. Our method (t-EVA) with 128 time-steps outperforms conventional, and MuViLab \cite{muvilab-github} methods with labeling 769 minutes of video in 21 minutes. Using more consecutive frames increases annotation speed.}
\begin{tabular*}{\columnwidth}{@{\extracolsep{\fill}}cccccc@{}}
\toprule
\textbf{} & \textbf{Conventional} & \textbf{MuViLab} & \textbf{t-EVA-32} & \textbf{t-EVA-64} & \textbf{t-EVA-128} \\ \midrule
\textbf{Time Gain}   & 3 x              & 4.5 x            & 18 x             & 24 x             & 36 x              \\ \bottomrule
\end{tabular*}
\label{tbl:time-comp}
\end{table}

First, we choose ActivityNet subset-1 with a total duration of 769 minutes. 
We annotated 30 minutes of videos using MuViLab and Conventional methods and extrapolated the result to match the total duration of ActivityNet subset-1. Additionally, the entire subset-1 is annotated using different variants of t-EVA, and we compare the annotation speed of all these methods (Table \ref{tbl:time-comp}).
%Table~\ref{tbl:time-comp} shows the result of the comparison between the conventional labeling and MuViLab versus our method in different time-steps. 
The results show that labeling 769 minutes of video takes approximately 42 minutes with the t-EVA-32 method. t-EVA-32 outperforms both conventional and MuViLab methods on ActivityNet subset-1 in annotation speed by a large margin by respectively 4 to 6 times faster. With t-EVA-64 and 128, time gain can reach respectively 24 and 36 times more.
Conventional annotation and MuViLab do not take advantage of the temporal dimension of videos for annotation. Nevertheless, our method exploits the spatio-temporal features and places similar actions near each other in the t-SNE plot for the oracle to annotate the actions.

%our method can achieve almost the same test accuracy on video classification task when trained on the ground-truth label, but much faster.

% Please add the following required packages to your document preamble:
% \usepackage{booktabs}

%%%%%%%%%%%%%%%%%%%%%%%%%%%%%%%%%
\begin{figure}[t!]
\centering
\includegraphics[width=0.7\columnwidth]{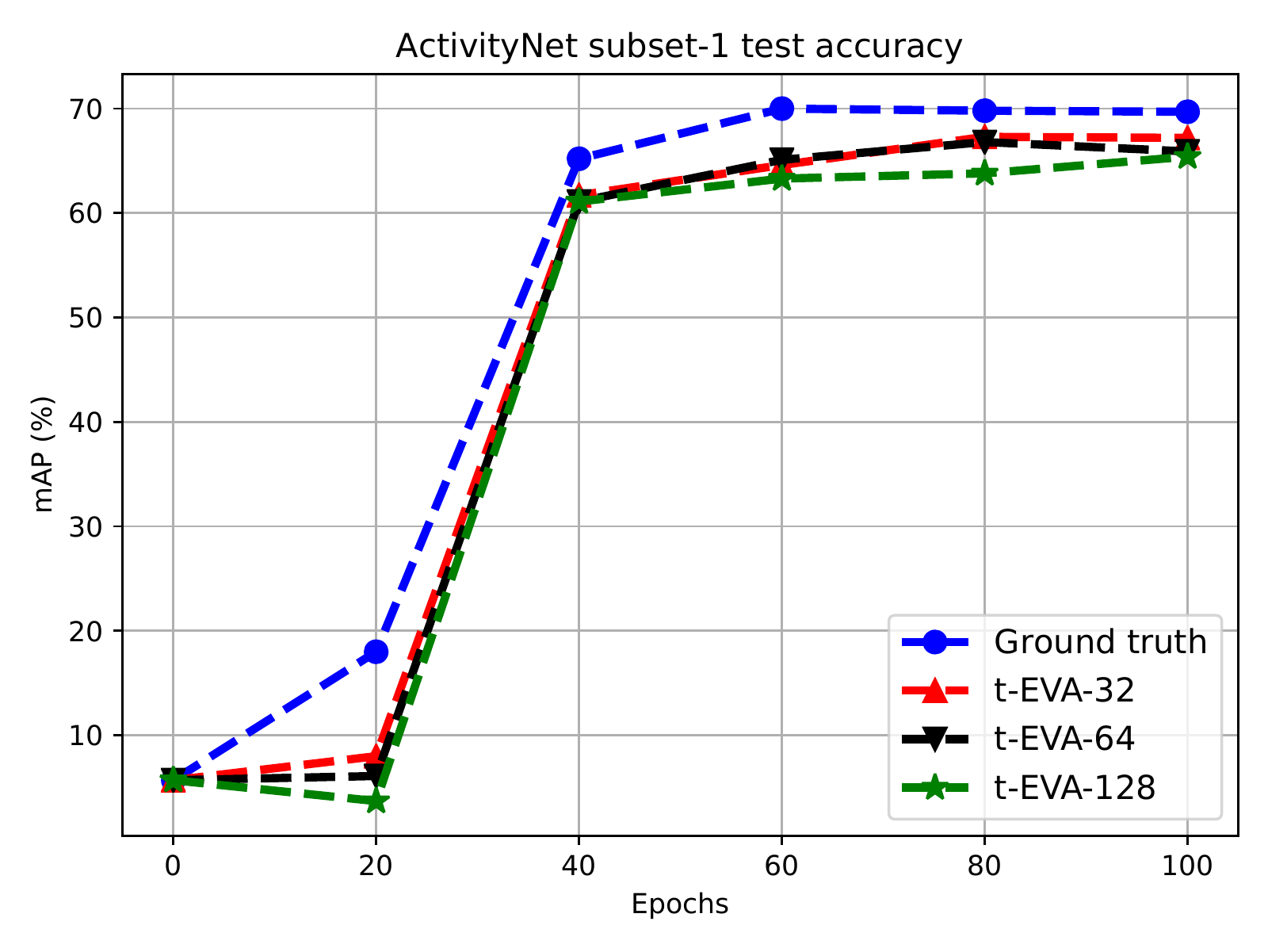}
\caption{Comparison of video classification performance in the form of mAP (\%) between fine-tuning the 3D ConvNet on ground truth label versus fine-tuning with our annotation acquired using different time-steps (TS). Fine-tuning the 3D ConvNet on the annotation generated by our method can achieve comparable video classification accuracy to the ground truth.}
\label{fig:different-time-steps}
\end{figure}
%%%%%%%%%%%%%%%%%%%%%%%%%%%%%%%%%

% \noindent{\textbf{Increasing Annotation Speed.} We further investigate increasing the time-steps in each clip to increase the annotation speed.
% One way to increase the annotation speed is by putting more videos on the screen for the oracle to annotate. However, since ActivityNet videos on average have 30 frames per second (FPS), every 32 time-steps that we sample represent almost 1 second $(\sim \frac{32}{30})$ of video. Putting all of the 407 videos (13 hours) overflows the screen with the frames and makes the annotation harder for the oracle. One way to prevent overflowing the figures with thousands of frames is to increase the time-steps for sampling frames for each clip to the point that the network can still preserve the temporal coherency between the clips. This way we can show all of the videos on the 2D plot with less number of points.}

%The experiments indicate that it is possible to increase the clip time-steps to 128 frames without losing much of the temporal coherency between the clips of the videos. 128 time-steps are almost equal to 4 seconds $(\sim \frac{128}{30})$ of video, which is almost the average length of a mid-range action of ActivityNet.

%Figure \ref{fig:different-time-steps} shows the test accuracy of 3D Resnet-34 fine-tuned with ground truth labels, fine-tuned with labels provided by the oracle using our method with three different time-steps. 

We also evaluate the performance of the network on the test set of ActivityNet subset-1. In Figure~\ref{fig:different-time-steps}, we compare the classification performance of the networks: (i) fine-tuned with original ground truth labels and (ii) fine-tuned by using newly annotated videos by 32, 64, and 128 time-steps.
%\noindent{Figure \ref{fig:different-time-steps} shows the video classification accuracy on the test set.
Annotating the videos with t-EVA method can achieve a classification performance of 67.2\% with 32-TS, 65.9\% with 64-TS, and 65.4\% with 128-TS, which is comparable to the training with ground truth labels (blue) by 69.7\% mAP.

Table \ref{tbl:map} shows the speed-accuracy trade off between t-EVA and ground-truth annotation.
When the original ground truth labels are used for fine-tuning the network, we obtain 69.7\% of mAP. 407 videos can be labeled in 42 minutes with t-EVA-32 by losing only 2.5\% of performance in comparison to using ground truth labels. When the time-steps are increased as 64 and 128, the annotation speed decreases respectively to 31 and 21 minutes, yet the classification performance also reduces by 3.8\% and 4.3\%.
Using 128 time-steps (t-EVA-128) reduces test accuracy while increasing the annotation speed. The decrease in accuracy compared to the 32-TS version is expected since the annotation is more prone to noise when the time-step is increased to 128 frames. With 128-TS for each clip, every point in the scatter plot represents 4 seconds of the video while it represents 1 second in the 32-TS version. Namely, labeling points wrongly in the 128 version (t-EVA-128) brings more significant consequences in the fine-tuning process. However, Table \ref{tbl:map} indicates that using 128-TS (t-EVA-128) compared to the 32-TS (t-EVA-32) increases the annotation speed twice while the mAP score decreases less than 2\%.

%Moreover, Table \ref{tbl:map} shows that incremental annotation and fine-tuning with 32 time-steps (32-TS) gives a very close mAP to the ground truth accuracy. Using 128 time-steps (128-TS) reduces test accuracy while increasing the annotation speed. The decrease in accuracy compared to the 32-TS version is expected since the annotation is more prone to noise when time-step is increased to 128 frames. With 128-TS for each clip, every point in the scatter plot represents 4 seconds of the video while it represents 1 second in the 32-TS version. Which means miss-labeling points in the 128 version comes with more significant consequences in the fine-tuning process. However, Table \ref{tbl:map} shows using 128-TS compared to the 32-TS, it increases the annotation speed twice while the mAP decreases less than 2\%.

\begin{table}[h!]
\centering
\caption{Comparison of video classification performance (mAP) and ToA (time of annotation) on ActivityNet subset-1. This subset contains 407 videos in about 13 hours of video. 
Our method in 32 time-steps (t-EVA-32) and 128 time-steps (t-EVA-128) achieves comparable test accuracy to the ground truth accuracy and requires a much shorter time to annotate. There is a trade-off between annotation speed and performance.}
\begin{tabular*}{\columnwidth}{@{\extracolsep{\fill}}ccccc@{}}
\toprule
\textbf{Method}          & \textbf{GT} & \textbf{t-EVA-32} & \textbf{t-EVA-64} & \textbf{t-EVA-128} \\ \midrule
\textbf{mAP}             & 69.7 \%     & 67.2 \%        & 65.9 \%        & 65.4 \%         \\
\textbf{ToA (minutes)}   & -           & 42             & 31             & 21              \\ \bottomrule
\end{tabular*}
\label{tbl:map}
\end{table}

\subsection{\textbf{Generalization}}
To further demonstrate the generalization of our method, we conduct the same annotation experiment on a more challenging subset of ActivityNet (v1.3) with 15 classes and a subset of Sports-1M \cite{sport1mCVPR14} with 15 random classes.

\textbf{ActivityNet (v1.3) Subset-2.} Subset 2 of ActivityNet (v1.3) contains 637 training videos and 311 test videos. The first iteration of features is extracted from the 637 training videos and is annotated in 15 minutes by the oracle using t-EVA. After 20 epochs of fine-tuning, the new features are extracted, and the labels are fine-tuned again by the oracle. After this stage, the network is fine-tuned for 80 epochs. After fine-tuning for 100 epochs, our method reaches a test accuracy of 66.4\%, while the training with ground-truth labels achieves an accuracy of 68.3\% on the video classification task. 

% To emulate the annotation speed in form of local homogeneity, we report KNN accuracy. 
The 4-NN accuracy of the final features is 92.4\%, which shows the quality of the extracted features is sufficient for the oracle to annotate. t-EVA can also perform well on the ActivityNet subset-2.
The fact validates that our method can also generalize on a more challenging subset of ActivityNet.

\begin{comment}
%%%%%%%%%%%%%our%%%%%%%%%%%%%%%%%%%%
\begin{figure}[h!]
\centering
\includegraphics[width=\columnwidth]{Figures/15class.png}
\caption{is this necessary}
\label{fig:15class}
\end{figure}
%%%%%%%%%%%%%%%%%%%%%%%%%%%%%%%%%
\end{comment}

\begin{figure}[t]
\centering
\includegraphics[width=0.8\textwidth]{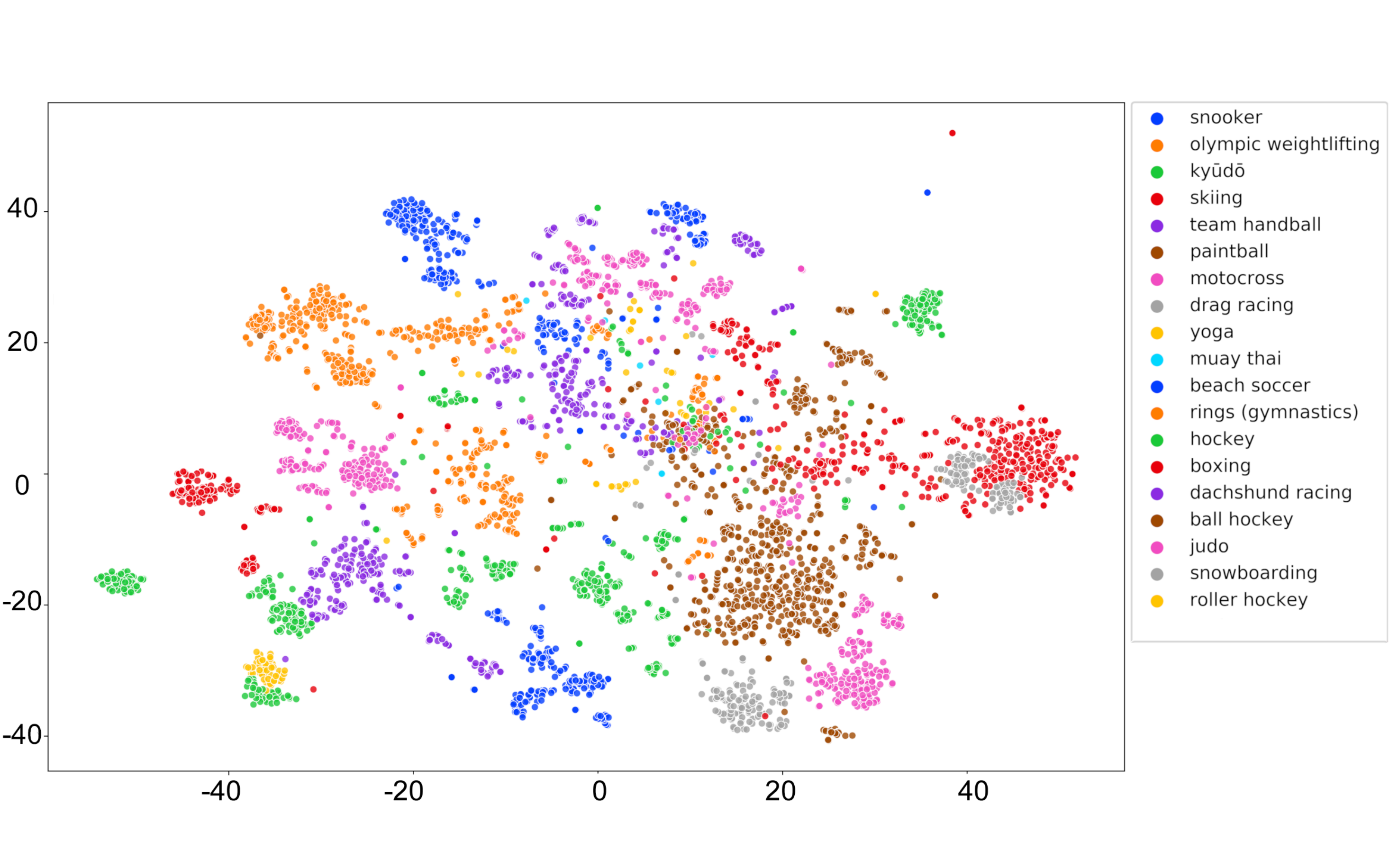}
\caption{t-SNE projection of extracted features from 200 videos from the Sports-1M \cite{sport1mCVPR14} dataset with ground truth labels as colors. 200 videos are from 15 random classes; however, some videos contain more than one activity class. The 4-NN accuracy, which emulates the quality of the projection through measuring local homogeneity, is 92.3$\%$, indicating such a figure is annotate-able by the oracle.}
\label{fig:sport-1m}
\end{figure}

\textbf{Sports-1M}. We further validate our method on a subset of Sports-1M \cite{sport1mCVPR14} dataset with 15 random classes. We randomly sample 200 videos ($\sim$860 minutes) from the total 702 videos available in the 15 classes. The features are extracted from 200 videos, and ground truth labels of the two-dimensional features can be seen in Figure \ref{fig:sport-1m}. Using 4-NN, we obtain an accuracy of 92.3\%, which shows the features can be annotated based on similarity. Using our method, we were able to annotate 860 minutes of video in 28 minutes, giving us a time gain of 30.7. 
%Similar to the ActivityNet subsets, 
t-EVA indicates an extensive time gain on the Sports-1M dataset. 

%%%%%%%%%%%%%%%%%%%%%%%%%%%%%%%%%

\section{\textbf{Ablation Study}}
In this section, we conduct an ablation study to motivate our design choices in the following aspects: (i) dimensionality reduction method, (ii) t-SNE parameter selection, and (iii) 2D versus 3D backbone for feature extraction.

\subsection{\textbf{Dimensionality Reduction}}
\begin{figure*}[ht!]
\centering
\includegraphics[width=\textwidth]{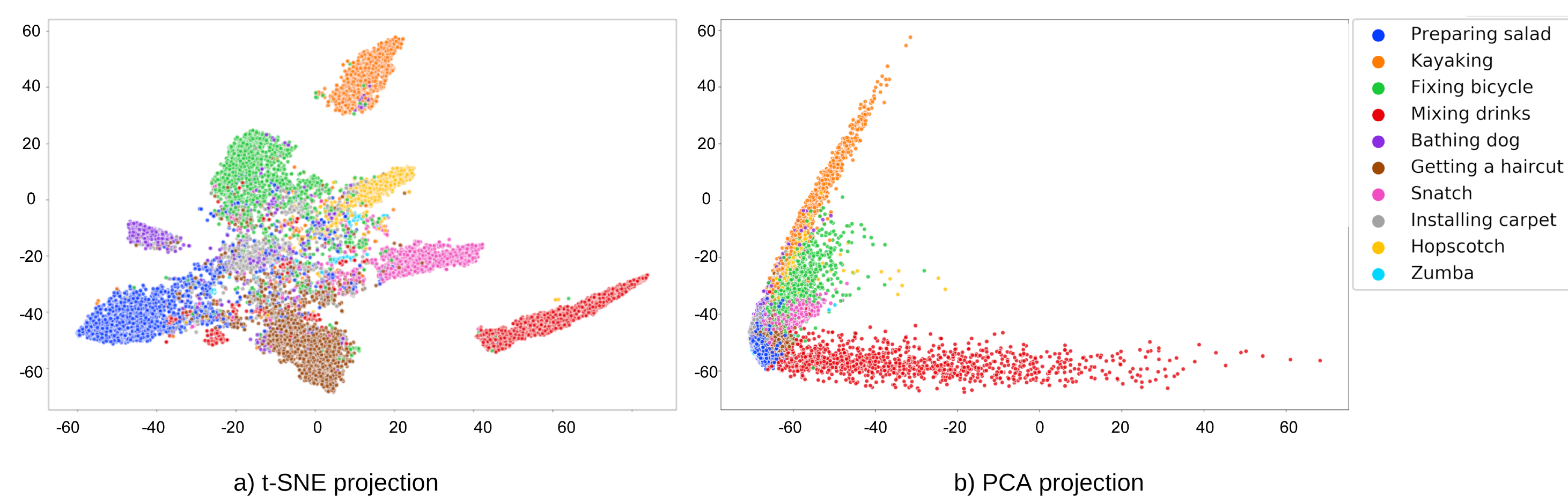}
\caption{Visual comparison of the projection quality of high-dimensional features to two dimensions using t-SNE (a) and PCA (b). 
%Linear DR methods such as 
PCA is unable to maintain the structure of the high-dimensional data in two dimensions.}
\label{fig:pca-tsne}
\end{figure*}

We investigate using PCA as a linear dimensionality method and t-SNE as a non-linear dimensionality method for visualizing the high-dimensional features in two dimensions. We use the extracted feature from the ActivityNet subset-1 with 407 videos. Figure \ref{fig:pca-tsne}-b shows that qualitatively PCA is not able to group similar features and separate unalike features from the videos in the transition to a lower dimension, making the annotation more difficult. However, Figure \ref{fig:pca-tsne}-a indicates that t-SNE projection can maintain the local structure of each class while separating the features from different classes. 
To report the quality of projection in quantitative measures, we use KNN with K=4. The 4-NN classification accuracy in Figure~\ref{fig:pca-tsne} for the t-SNE projection is 80.6\%, and for the PCA projection is 58.2\%. Therefore, PCA, a linear dimensionality method, cannot reduce the feature dimension while placing similar classes near each other.

% We further compare our result with UMAP \cite{UMAP-lel2018umap} versus Barnes-Hut t-SNE \cite{barnes-haut-maaten2013barneshutsne} for ActivityNet subset-1.  It can be seen in Table \ref{tbl:umap}, UMAP offers a faster computation time than Barnes-Hut t-SNE which is not an issue since faster implementations of t-SNE are available such as \cite{gpgpu-PTMHLLEV20, tsne-cuda-chan2018tsnecuda}.

% Please add the following required packages to your document preamble:
% \usepackage{booktabs}
% \begin{table}[]
% \centering
% \caption{Comparison of quality of projection and execution time (Exec Time) of t-SNE versus UMAP through measuring 4-NN accuracy to estimate local homogeneity. UMAP has a faster execution time than BH-t-SNE but t-SNE can give a better projection with higher local homogeneity which can speed up the annotation process.}
% \begin{tabular*}{\columnwidth}{@{\extracolsep{\fill}}ccc@{}}
% \toprule
%               & \textbf{4-NN} & \textbf{Exec Time} \\ \midrule
% \textbf{t-SNE} & 94\%          & 160                \\
% \textbf{UMAP}  & 83.3\%        & 40                 \\ \bottomrule
% \end{tabular*}
% \label{tbl:umap}
% \end{table}

%%%%%%%%%%%%%%%%%%%%%%%%%%%%%%%%%%
%%%%%%%%%%%%%%%%%%%%%%%%%%%%%%%%%

\subsection{\textbf{t-SNE Parameters}}
We investigate using different perplexity parameters for the t-SNE projection. \cite{tsne-vandermaten-7b54165e73a3424b8820136bcf61ca89} recommend using perplexity parameter between [5-50], however larger and denser datasets requires relatively higher perplexity. With low perplexity, the local structure of data in each video dominates the action grouping from multiple video \cite{tsne-pitfall-Wattenberg2016HowTU}, but our goal is to group multiple actions from different videos. To emulate the t-SNE projection quality for the annotation, we report homogeneity and completeness scores with different perplexities in Table \ref{tbl:homogenity}. Perplexity 30 shows the highest homogeneity and completeness scores, in other words, t-SNE projection with perplexity 30 can separate the classes better than projecting with the other perplexity parameters. Therefore, using t-SNE with perplexity 30 makes the group labeling process easier for the oracle.

% Please add the following required packages to your document preamble:
% \usepackage{booktabs}
\begin{table}[]
\centering
\caption{Comparison of homogeneity and completeness scores as a measure to emulate the quality of t-SNE projection on a global-level. Higher homogeneity means all the points in a cluster belong to the same class. Higher completeness means all the points belonging to a class are in the same cluster. t-SNE perplexity parameter as 30 gives the highest homogeneity and completeness score.}
\begin{tabular*}{\columnwidth}{@{\extracolsep{\fill}}ccccccc@{}}
\toprule
                      & \textbf{px-5} & \textbf{px-15} & \textbf{px-30} & \textbf{px-50} & \textbf{px-100} & \textbf{px-120} \\ \midrule
\textbf{Homogeneity}  & 44.7\%         & 58.7\%          & 62.5\%          & 61.3\%          & 61.7\%           & 61.5\%           \\
\textbf{Completeness} & 42.5\%         & 56.1\%          & 60\%            & 58.5\%          & 59\%             & 58.8\%           \\ \bottomrule
\end{tabular*}
\label{tbl:homogenity}
\end{table}

\subsection{\textbf{2D-3D Comparison}}
\begin{figure}[ht]
\includegraphics[width=\textwidth]{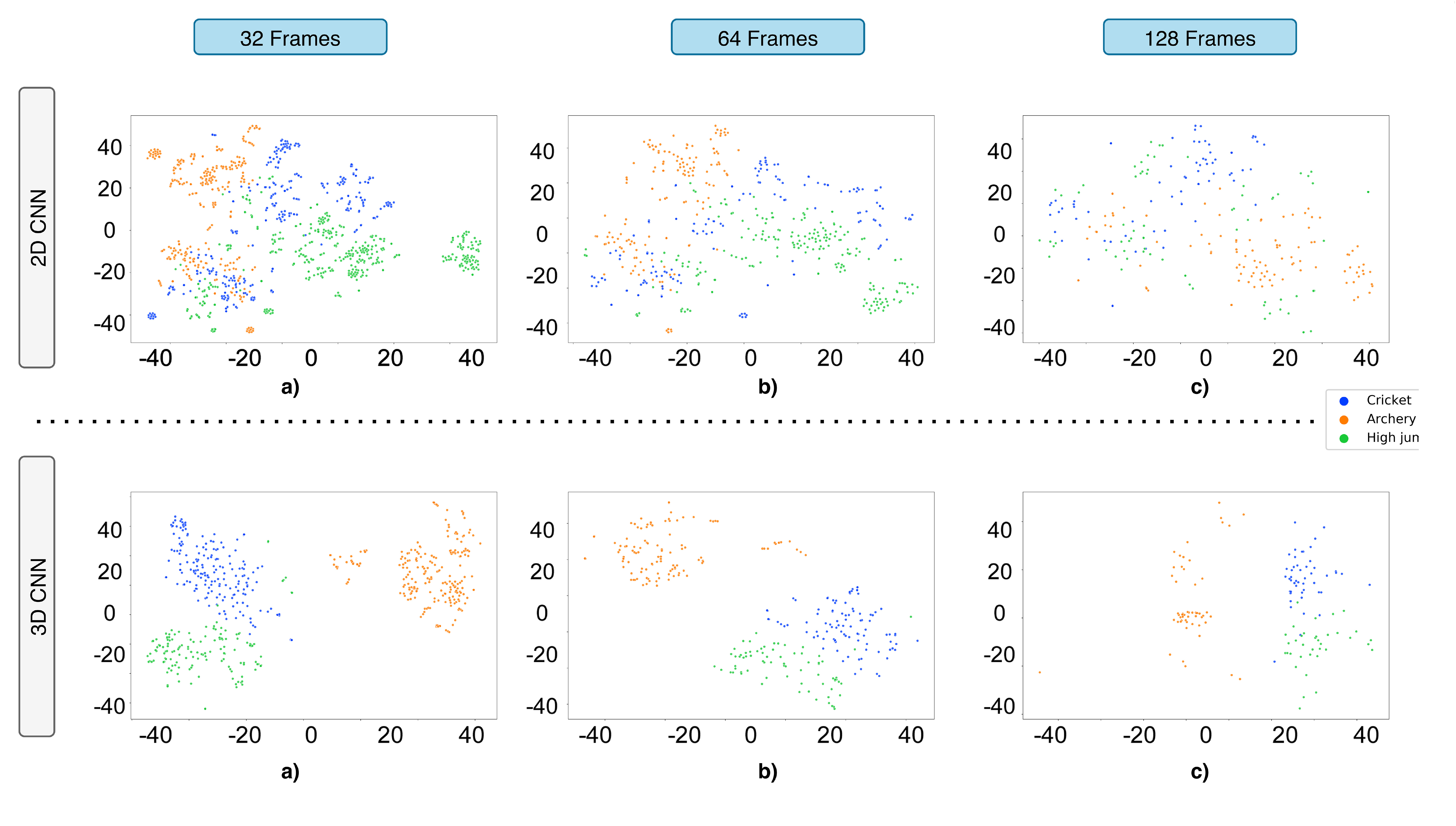}
\caption{Comparison of t-SNE projection of extracted features from a 2D CNN versus a 3D ConvNet for videos from 3 action classes of ActivityNet dataset \cite{activitynet-refrence}. Increasing the time-steps for sampling clips from the videos causes the 2D CNN to lose the spatial information of the clips. However, the features from the 3D ConvNet can maintain the coherency between the clips.} 
%as they take time into account in the convolution kernels.}
\label{fig:2d-3d-comp}
\end{figure}

We investigate replacing the 3D ConvNet with a 2D CNN to compare the quality of the feature embedding. For 3D ConvNet, 3D ResNet-34 pre-trained on Kinetics \cite{kinetics-DBLP:journals/corr/KayCSZHVVGBNSZ17} and for the 2D CNN ResNet-50 pre-trained on Kinetics \cite{kinetics-DBLP:journals/corr/KayCSZHVVGBNSZ17} are used. We chose Resnet-50 instead of Resnet-34 for the 2D CNN because the Kinetics pre-trained weights were only available for ResNet-50. 
To experiment, we sample every 32 consecutive frames (time-steps) as a clip in the 3D ConvNet, and for the 2D CNN, we choose one frame for every 32 frames to represent that specific window. The experiment is done on the subset-1 of the Activity-Net dataset with 10 classes. It can be seen in Figure~\ref{fig:2d-3d-comp} that we start the experiments with 32 time-steps. 
%Later we gradually increase the time-steps to the point that either we lose temporal information or we run out of data-points (whichever comes first). 
With 32 time-steps, we can see the 2D CNN can capture the same action in different videos but can not place them together as well as the 3D ConvNet. Therefore, the colors representing the classes are better gathered nearby in the 3D ConvNet, making the annotation process faster than the 2D CNN projection.
Moreover, by increasing the time-steps for frame sampling, the 2D CNN, even with deeper architecture, starts losing the temporal coherency between the data-points because 2D CNN only focuses on the spatial information between the frames. Focusing only on spatial information can still work in lower time-steps (32-TS) since the frames from the same action contain similar spatial information. However, using spatial information alone becomes problematic in higher time-steps as increasing the time-steps reduces the spatial similarity between the frames.

To evaluate our findings quantitatively, we use K-NN accuracy as a quantitative emulation for the quality of features for annotation.
Table \ref{tbl:knn_2d_3d} shows that increasing the number of frames in the clips degrades the 4-NN accuracy of 2D CNN dramatically from 93\% to 75\%. However, 3D CNN only loses around 5\% from 32 time steps to 128. The local homogeneity decreases more drastically in 2D CNNs compared to 3D CNNs, which makes annotation more difficult for the oracle. In other words, the 2D CNN alone can not maintain the temporal structure of the data in higher time-steps. Thus, in the t-EVA method, 3D features are extracted to use for group labeling.
%%%%%%%%%%%%%%%%%%%%%%%%%%%%%%%%%%
%%%%%%%%%%%%%%%%%%%%%%%%%%%%%%%%%
%%%%%%%%%%%%%%%%%%%%%%%%%%%%%%%
\begin{table}[h!]
\centering
\caption{Comparison of 4-NN accuracy of extracted features from a 2D CNN (ResNet-50) and a 3D ConvNet (3D ResNet-34) on subset-1 of ActivityNet \cite{activitynet-refrence}. Increasing time-steps cause the 2D CNN to lose the spatial similarity between the frames and fail to group them in the t-SNE plot, while the 3D ConvNet can still group similar actions even in higher time-steps.}
\begin{tabular*}{\columnwidth}{@{\extracolsep{\fill}}llll@{}}
\toprule
                & \textbf{32-TS} & \textbf{64-TS} & \textbf{128-TS} \\ \midrule
\textbf{2D CNN} & 93.1 \%        & 89.3 \%        & 74.6 \%         \\
\textbf{3D CNN} & 100 \%         & 97.6 \%        & 95.2 \%         \\ \bottomrule
\end{tabular*}
\label{tbl:knn_2d_3d}
\end{table}

% \newpage
\section{\textbf{Conclusion}}
This paper introduced a smart annotation tool, t-EVA, for helping the oracle to group label videos based on their latent space feature similarity in two-dimensional space.
Our experiments on subsets of large-scale datasets shows that t-EVA can be useful in annotating large-scale video datasets, especially if the annotation budget and time are limited. Our method can outperform the conventional annotation method, and MuViLab \cite{muvilab-github} time-wise in the order of magnitude with a minor drop in the video classification accuracy. 
Besides, t-EVA is a modular tool, and its components can be easily replaced by other methods. To illustrate, 3D ResNet can be changed to another feature extractor.

t-EVA method has a trade-off between annotation speed and network performance. Increasing time steps can reduce the annotation time; however, the network's accuracy may also decrease. 

t-EVA can be sensitive to the initial state of the feature extractor. If the feature extractor can not separate classes well, it can take a longer time to annotate the videos initially. %Therefore, other feature extractors need to be used.
After fine-tuning the network with new labels for a few epochs, the labeling time can reduce again. Besides, putting more video frames in the t-SNE plot can overflow the screen and make the annotation process harder for the oracle.

\newpage
\bibliography{main}
\bibliographystyle{splncs04}
%
% \begin{thebibliography}{8}
% \bibitem{ref_article1}
% Author, F.: Article title. Journal \textbf{2}(5), 99--110 (2016)

% \bibitem{ref_lncs1}
% Author, F., Author, S.: Title of a proceedings paper. In: Editor,
% F., Editor, S. (eds.) CONFERENCE 2016, LNCS, vol. 9999, pp. 1--13.
% Springer, Heidelberg (2016). \doi{10.10007/1234567890}

% \bibitem{ref_book1}
% Author, F., Author, S., Author, T.: Book title. 2nd edn. Publisher,
% Location (1999)

% \bibitem{ref_proc1}
% Author, A.-B.: Contribution title. In: 9th International Proceedings
% on Proceedings, pp. 1--2. Publisher, Location (2010)

% \bibitem{ref_url1}
% LNCS Homepage, \url{http://www.springer.com/lncs}. Last accessed 4
% Oct 2017
% \end{thebibliography}
\end{document}